\documentclass[10pt,twocolumn,letterpaper]{article}

\usepackage[pagenumbers]{cvpr} 

\usepackage{booktabs}

\definecolor{cvprblue}{rgb}{0.21,0.49,0.74}
\usepackage[pagebackref,breaklinks,colorlinks,allcolors=cvprblue]{hyperref}
\usepackage{multirow}
\usepackage{colortbl}
\usepackage{fancyhdr}
\usepackage{amssymb}
\usepackage{bbm}
\usepackage{pifont}
\usepackage[accsupp]{axessibility}

\newcommand{\mysection}[1]{\vspace{0.05pt}\noindent\textbf{#1}}

\def\paperID{54} 
\def\confName{CVPR}
\def\confYear{2025}

\title{Action Anticipation from SoccerNet Football Video Broadcasts}

\author{Mohamad Dalal$^{1,}$\thanks{Equal contribution.}\hspace{0.7cm} 
Artur Xarles$^{2, 3,\ast}$\hspace{0.7cm} 
Anthony Cioppa$^{4}$\hspace{0.7cm} 
Silvio Giancola$^{5}$\\
Marc Van Droogenbroeck$^{4}$\hspace{0.7cm} 
Bernard Ghanem$^{5}$\hspace{0.7cm} 
Albert Clapés$^{2, 3}$\hspace{0.7cm} 
Sergio Escalera$^{1, 2, 3}$\\
Thomas B. Moeslund$^{1}$\\
$^{1}${\small Aalborg University}\hspace{0.7cm} 
$^{2}${\small Universitat de Barcelona}\hspace{0.7cm} 
$^{3}${\small Computer Vision Center}\hspace{0.7cm} 
$^{4}${\small University of Li{\`e}ge}\hspace{0.7cm} 
$^{5}${\small KAUST}\\
}

\begin{document}
\maketitle


\begin{abstract}
Artificial intelligence has revolutionized the way we analyze sports videos, whether to understand the actions of games in long untrimmed videos or to anticipate the player's motion in future frames.
Despite these efforts, little attention has been given to anticipating game actions before they occur.
In this work, we introduce the task of action anticipation for football broadcast videos, which consists in predicting future actions in unobserved future frames, within a five- or ten-second anticipation window.
To benchmark this task, we release a new dataset, namely the SoccerNet Ball Action Anticipation dataset, based on SoccerNet Ball Action Spotting. 
Additionally, we propose a Football Action ANticipation TRAnsformer (FAANTRA), a baseline method that adapts FUTR, a state-of-the-art action anticipation model, to predict ball-related actions.
To evaluate action anticipation, we introduce new metrics, including mAP@$\delta$, which evaluates the temporal precision of predicted future actions, as well as mAP@$\infty$, which evaluates their occurrence within the anticipation window. We also conduct extensive ablation studies to examine the impact of various task settings, input configurations, and model architectures. 
Experimental results highlight both the feasibility and challenges of action anticipation in football videos, providing valuable insights into the design of predictive models for sports analytics. 
By forecasting actions before they unfold, our work will enable applications in automated broadcasting, tactical analysis, and player decision-making. 
Our dataset and code are publicly available at \href{https://github.com/MohamadDalal/FAANTRA}{https://github.com/MohamadDalal/FAANTRA}.
\end{abstract}

\begin{figure}[t]
    \centering
    \includegraphics[width=1.0\linewidth]{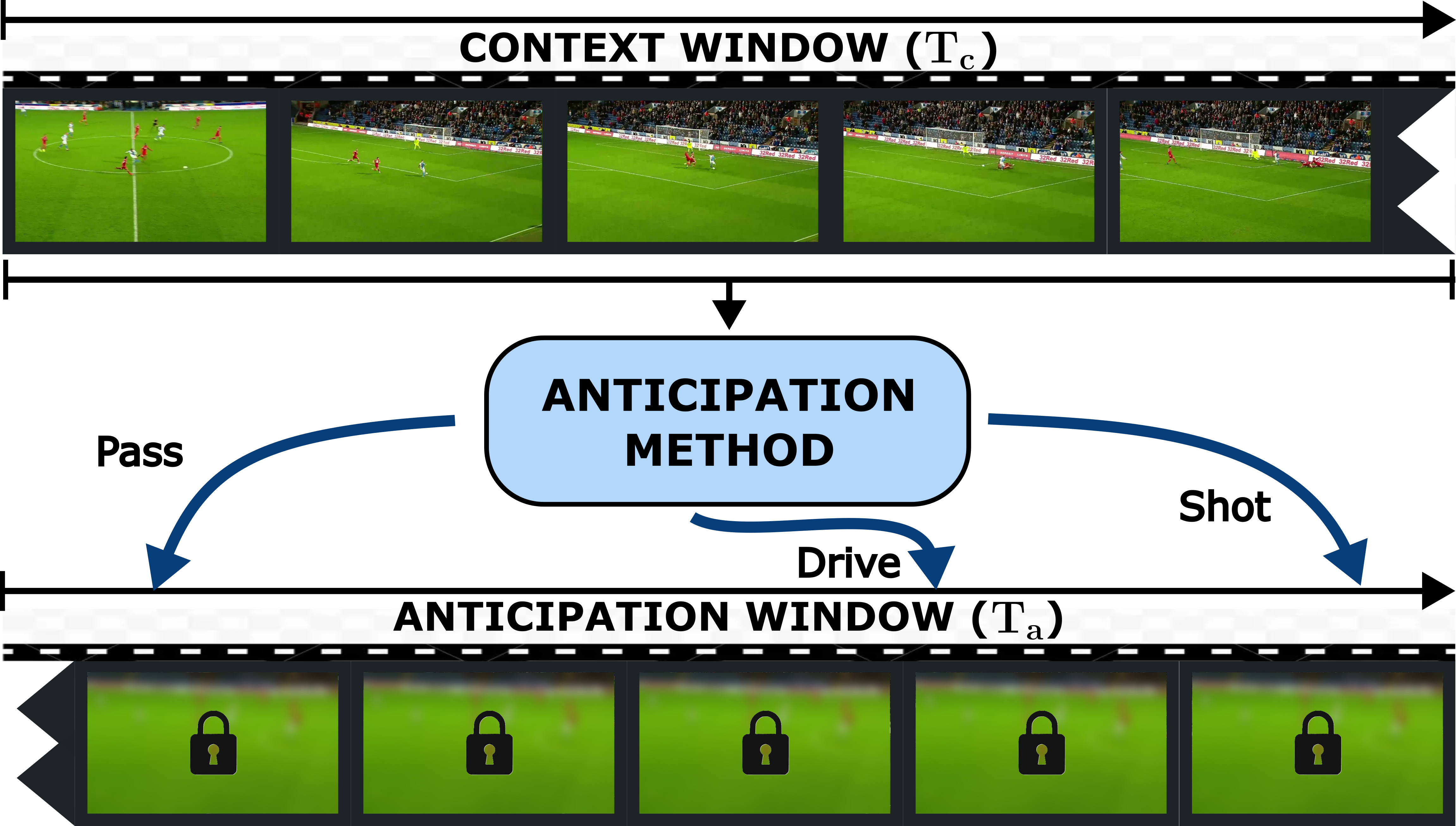}
    \caption{
    \textbf{Overview of our new action anticipation task for sports.} 
    Action anticipation aims to predict and temporally localize future actions in an \emph{anticipation window} of $T_a$ seconds using information from a preceding observed \emph{context window} of $T_c$ seconds. 
    Unlike action spotting, where models can access the entire video sequence to detect actions, action anticipation requires predicting future events without access to future frames.
    }
    \label{fig:pulling}
\end{figure}


\section{Introduction}

Artificial intelligence and computer vision have significantly advanced sports analytics, improving player tracking, tactical evaluation, and event detection. These advancements support applications ranging from real-time match analysis to automated content generation. Much progress has been made in \textit{action recognition}~\cite{wu2022survey}, which classifies actions, and \textit{action spotting}~\cite{Giancola2024Deep}, which localizes them in time. Yet, \textit{action anticipation}, which predicts actions before they occur, remains largely unexplored.  
Anticipation is a key aspect in football, and integrating this predictive capability into AI can enhance game understanding and enable applications such as:
(1) automated broadcasting, where cameras proactively adjust based on anticipated actions,
(2) team tactical analysis, where coaches gain insights into possible future plays, and
(3) player decision-support systems, which help athletes anticipate opponents' moves and optimize their positioning.

Most prior work on future game dynamics prediction has focused on \textit{trajectory forecasting}, for player or ball movement~\cite{goka2024and, wei2013predicting, chang2023will}, rather than anticipating actions such as passes or shots. Unlike action spotting, which detects actions using full video sequences, action anticipation requires reasoning about upcoming actions using only past observations. This paradigm shift introduces additional challenges, requiring models to recognize patterns in past actions and infer likely outcomes.

This work introduces the first structured benchmark for action anticipation in football by presenting the SoccerNet Ball Action Anticipation dataset, an adaptation of previous versions of the SoccerNet dataset~\cite{giancola2018soccernet, Deliege2021SoccerNetv2, Cioppa2024SoccerNet2023Challenge}. As an initial solution to solve the task, we propose the Football Action ANticipation TRAnsformer (FAANTRA) model, a baseline that adapts FUTR~\cite{gong2022future}, a state-of-the-art action anticipation model, to predict ball-related actions within five- and ten-second future windows, using only past frames. In line with current trends in action spotting~\cite{xarles2024t, Hong2022Spotting}, we employ an efficient video feature extractor that we train in an end-to-end approach. To evaluate performance, we propose new anticipation metrics, including mAP@$\delta$ for temporal evaluation and mAP@$\infty$ for action occurrence within the anticipation window.  
Additionally, we conduct extensive ablation studies to analyze the impact of various task settings, input configurations, and model architectures. These studies highlight the challenges of the task while also underscoring the importance of key modeling choices.

\mysection{Contributions.} We summarize our contributions as follows:
    \textbf{(i)} We introduce the first structured benchmark for action anticipation in sports, formalizing the task with a novel SoccerNet Ball Action Anticipation dataset and pertinent metrics to evaluate that task.
    \textbf{(ii)} We propose FAANTRA, the first baseline model specifically designed for action anticipation in football broadcasts.  
    \textbf{(iii)} We conduct extensive ablation studies, providing insights into the challenges of the anticipation task as well as effective model design.

\section{Related work}
\label{sec:related}

\mysection{Action spotting.}
Given a long, untrimmed video stream, the task of action spotting consists in identifying and precisely localizing actions of interest in time~\cite{giancola2018soccernet,Seweryn2023Survey, Giancola2024Deep}.
Automatically extracting actions is key for many sports applications, including generating game statistics~\cite{Deliege2021SoccerNetv2}, supporting video analysts in coaching or referees during games~\cite{Held2023VARS,Held2024XVARS}, or customizing highlights based on the viewer's preferences~\cite{Valand2021AIBased, Valand2021Automated}.
The SoccerNet challenges~\cite{Giancola2022SoccerNet, Cioppa2024SoccerNet2023Challenge, Cioppa2024SoccerNet2024Challenge-arxiv} have led to the development of numerous methods, which can be categorized into two training paradigms: \emph{feature-based approaches}~\cite{Giancola2021Temporally,Cioppa2020AContextaware,Zhou2021Feature-arxiv,Soares2022Temporally,Tomei2021RMSNet}, which rely on pre-extracted features and train only the spotting head, and \emph{end-to-end approaches}~\cite{Hong2022Spotting,Baikulov2023Solution,Denize2024COMEDIAN,Xarles2023ASTRA}, which simultaneously train both the backbone and the spotting head.
Yet, all these methods require access to the complete video stream containing the actions to spot them, including past, present, and future contextual frames. 
In this work, we propose to tackle the task of action anticipation, in which the methods only have access to past context to predict future actions. 
Particularly, we follow current trends in action spotting by proposing an end-to-end approach to anticipation, leveraging an efficient feature extractor from T-DEED~\cite{xarles2024t}, the state-of-the-art (SOTA) method for action spotting.

\mysection{Action anticipation.} This task involves predicting future actions based solely on the observation of past context. It can be divided into two main types: short-term and long-term action anticipation. In \textit{short-term action anticipation}~\cite{xu2021long, zhao2022real, girase2023latency, furnari2020rolling, zhong2023anticipative}, models predict actions within a small fixed future window, typically ranging from $1$ to $5$ seconds, which defines the action class. This is usually done by dividing the window into intervals and predicting the action occurring at the end of each one. In contrast, \textit{long-term action anticipation}~\cite{abu2018will, gong2022future, nawhal2022rethinking, zhong2023diffant, zhang2024object, Zhao2024AntGPT} aims to predict the sequence of future actions along with their durations, and may look up to several minutes ahead. Our proposed action anticipation task in football aligns with the short-term setting, anticipating actions within fixed five- or ten-second windows. However, rather than localizing activities at the end of predefined intervals, the objective is to localize the exact timestep of each anticipated action.

Most action anticipation methods rely on pre-extracted features to address the task~\cite{zhong2023anticipative, nagarajan2020ego, gammulle2019forecasting, abu2021long, mascaro2023intention}, while only a few tackle it in an end-to-end manner~\cite{girdhar2021anticipative, zhong2018unsupervised, li2022mvitv2,liang2019peeking}. 
To process information within the observed time window, these methods typically employ architectures designed to capture the sequential nature of videos, such as Recurrent Neural Networks (RNNs)~\cite{furnari2020rolling,liang2019peeking,sun2019relational, abu2018will, gammulle2019forecasting}. 
More recently, Transformers have become the preferred choice, as seen in methods like FUTR~\cite{gong2022future} and Anticipatr~\cite{nawhal2022rethinking}, both of which leverage a transformer encoder for this purpose. 
When generating anticipations, some methods~\cite{abu2018will, sener2020temporal, abu2021long, abu2019uncertainty} adopt an auto-regressive approach, where predictions are made sequentially. Non auto-regressive approaches, on the other hand, typically fall into one of two main categories: anchor-based and query-based approaches. Anchor-based approaches~\cite{Zhao2024AntGPT, xu2021long, zhao2022real} consider predefined temporal positions within the anticipation window, predicting the action occurring at each position. Most short-term anticipation methods fall into this category, as they focus on predicting actions within fixed temporal intervals. In contrast, query-based approaches~\cite{nawhal2022rethinking, gong2022future, zhong2023diffant} use a set of learnable representations that are not tied to specific temporal positions, instead predicting both the action and its precise localization. FAANTRA aligns with current trends by leveraging transformers while adopting a query-based approach, which our ablation studies show to be beneficial.

Action anticipation datasets are typically adapted from action localization or segmentation datasets, often focusing on daily activities, such as cooking~\cite{stein2013combining, kuehne2014language, li2018eye, damen2022rescaling}, or sourced from movie and TV-series datasets~\cite{de2016online, gu2018ava}. However, no dataset currently addresses action anticipation in sports. Therefore, our proposed SoccerNet Ball Action Anticipation dataset is the first \emph{video-based action} anticipation dataset focusing on sports, specifically football. Moreover, the fast-paced and uncertain nature of football adds an extra challenge compared to other datasets, where the sequence of actions tends to follow a more scripted pattern.

\mysection{Anticipation in sports.} While research on anticipation in sports exists, none of the current studies approach it from a video-based action anticipation perspective. Instead, some methods~\cite{felsen2017will, goka2024and} rely on ball and player tracking data along with metadata and are constrained to anticipating only the next action class. In contrast, other works~\cite{wei2013predicting, wei2015forecasting, chang2023will} focus on predicting the trajectories of the ball and players, which differs from our task of anticipating actions.

\section{Action anticipation in sports}
\label{sec:task}

\mysection{Task definition.}  
Given a set of $N$ observed, trimmed videos $\mathcal{V}=\{v^1, v^2, \dots, v^n, \dots, v^N\}$ with a supposed fixed length of $T_c$ seconds, the task of \textbf{\textit{action anticipation}} aims at predicting a set of future actions of interest $\mathcal{A}^n=\{a_1^n,a_2^n,\dots,a_k^n,\dots\,a_{|\mathcal{A}^n|}^n\}$ for each video $v^n$, where $|\mathcal{A}^n|$ is the total number of future actions of interest for video $n$, which may vary across different videos. More precisely, the goal is to only use the observed video $v^n$, also referred to as the \textit{context window}, to identify and localize all actions that will occur within its immediate unobserved future time window $w^{n}\in\mathcal{W}$, also called the \textit{anticipation window}, with a fixed length of $T_a$ seconds. All actions $a_k^n$ are defined by two features: (1) a class $c\in\{1,\dots,C\}$, $C$ being the total number of action classes of interest, and (2) a timestamp $t\in[0,T_a]$, indicating the exact moment the action will happen in the anticipation window. 

\mysection{Evaluation metrics.} 
To benchmark our action anticipation task, we adapt the mean Average Precision (mAP) metric, proposed by \citet{giancola2018soccernet} for Action Spotting (AS), to evaluate performance within the non-observed anticipation window. Specifically, the mAP for a given temporal tolerance $\delta$ (\ie, a prediction is considered correct if it falls within a window of $\delta/2$ seconds before or after the ground truth), denoted as mAP@$\delta$, is computed by averaging the Average Precision (AP) values across different action classes~\cite{Giancola2024Deep}. AP summarizes precision-recall points into a single value by estimating the area under the precision-recall curve.
To evaluate different aspects of model performance, we consider six variations of the mAP@$\delta$ metric with different $\delta$ tolerances: $\{1, 2, 3, 4, 5, \infty \}$. Smaller tolerances emphasize on precise localization of anticipated actions, while larger tolerances are more loose on minor localization errors. When $\delta=\infty$, the metric completely disregards localization and focuses only on the correct identification and counting of action classes. These variations provide a balanced evaluation of both localization accuracy and action identification. Finally, we report the average performance across all six variations.

\section{SoccerNet Ball Action Anticipation dataset}
\label{sec:dataset}

We introduce the SoccerNet Ball Action Anticipation dataset (SN-BAA), by adapting the SoccerNet Ball Action Spotting dataset (SN-BAS) for our novel action anticipation task. SN-BAS comprises untrimmed videos from nine professional football matches, annotated with the precise localization of $C=12$ different on-ball action types. 
The dataset is divided into four splits: four games for training, one for validation, two for testing, and two with hidden ground truth for challenge evaluation.

\begin{table}[t]
\resizebox{\linewidth}{!}{
\begin{tabular}{lccccccc}
\toprule
Dataset                                   & Year & Domain  & Spontaneity & Hours & Classes \\
\midrule
50Salads\cite{stein2013combining}         & 2013 & Cooking & Medium      & 4.5   & 17      \\
Breakfast\cite{kuehne2014language}        & 2014 & Cooking & Medium      & 77    & 48      \\
THUMOS14\cite{idrees2017thumos}           & 2014 & Web     & Medium      & 20    & 20      \\
Charades\cite{sigurdsson2016hollywood}    & 2016 & ADL     & Low         & 82.3  & 157     \\
TVSeries\cite{de2016online}               & 2016 & Movie   & Medium      & 16    & 30      \\
EGTEA Gaze+\cite{li2018eye}               & 2018 & Cooking & Medium      & 28    & 106     \\
EpicKitchens-100\cite{damen2022rescaling} & 2020 & Cooking & High        & 100   & 3807    \\
Ego4D\cite{grauman2022ego4d}              & 2022 & ADL     & High        & 243   & 4756    \\
Assembly101\cite{sener2022assembly101}    & 2022 & ADL     & Medium      & 513   & 1380    \\
\midrule
SN-BAA (ours)                             & 2025 & Football  & Very High   & 11.4  & 10      \\
\bottomrule
\end{tabular}
}
\caption{Comparison of video-based action anticipation datasets, including SN-BAA, with details on the year, domain, spontaneity level, total hours, and number of action classes, following~\citet{zhong2023survey}. ADL stands for Activities of Daily Living.}
\label{tab:datasets}
\end{table}

In adapting the dataset for the anticipation task, we preserve the original splits from SN-BAS. For the training and validation splits, we retain the original untrimmed videos, allowing different methods to process them freely based on their characteristics. To ensure standardized evaluation on the test and challenge splits, we clip the untrimmed videos into $30$ seconds video clips, using a sliding window with a stride of $T_a$ seconds.
This enables methods to work with context windows $T_c$ ranging from $0$ to $30$ seconds. Using a stride of $T_a$ seconds ensures that all actions are evaluated.
Regarding the action classes considered, we exclude goals and free-kicks from the SN-BAA dataset, reducing the set to $C=10$ classes. As detailed in the dataset analysis in the supplementary material, these classes are rare, with only six and two observations, respectively, in the test split. Such low frequency leads to an unstable metric, where a single correct or incorrect anticipation can cause large fluctuations in performance. Therefore, to ensure a more robust evaluation, we removed these classes from SN-BAA.

\begin{figure*}[t]
    \centering
    \includegraphics[width=1\linewidth]{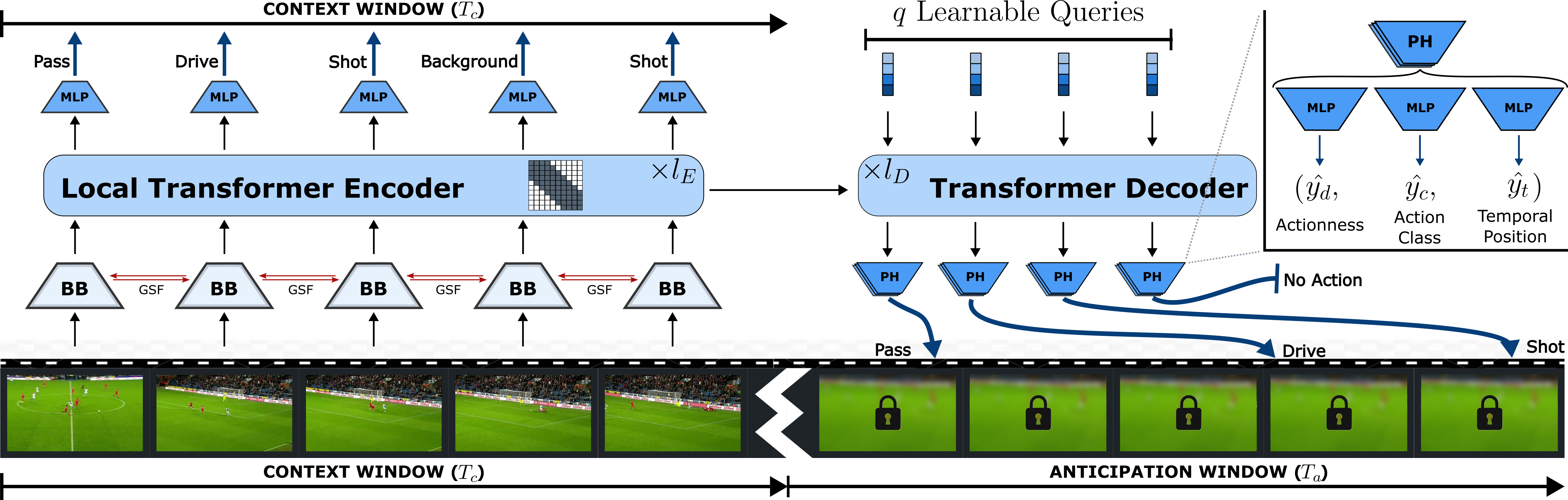}
    \caption{\textbf{FAANTRA Architecture Overview.} FAANTRA processes context video frames by extracting per-frame representations through a backbone (BB). These features are fed into a transformer encoder to capture temporal dependencies. A set of learnable queries, representing action predictions, are initialized in the transformer decoder and refined through multiple layers, leveraging information from the encoder. Each refined query is then processed by a prediction head (PH) to output three components representing the anticipated actions: action detection (\ie, actionness), action class, and temporal position.}
    \label{fig:enter-label}
\end{figure*}

Comparing the SN-BAA dataset with other common video-based action anticipation datasets in \cref{tab:datasets}, we find that our dataset stands out in terms of spontaneity (\ie, the unpredictability and suddenness of actions). The fast-paced and adversarial nature of football, where players attempt to hide their intentions from opponents, make the actions in our dataset less predictable and more challenging than in others. Additionally, SN-BAA is the first action anticipation dataset focused on sports, specifically in football. However, it has a moderate number of hours and classes compared to other datasets. Further analysis of the dataset statistics in the supplementary material reveals a total of $12433$ actions across the entire dataset (\ie, one action every $3.30$ seconds). The action classes follow a long-tail distribution, with passes and drives being the most frequent, while actions such as shots or successful tackles occur far less often. A similar pattern is observed when examining the distribution of actions within an anticipation window of $T_a = 5$ seconds, where the dataset contains an average of $1.5$ actions per window, with a maximum of $8$ actions.

\section{Football action anticipation transformer}
\label{sec:method}

\mysection{Method overview.} Our new method, namely Football Action ANticipation TRAnsformer (FAANTRA), builds on top of FUTR~\cite{gong2022future}, a SOTA method for action anticipation.
Specifically, our approach consists of three main components: a feature extractor, a transformer encoder-decoder, and a prediction head. Following current trends in Action Spotting (AS)~\cite{Hong2022Spotting}, we employ end-to-end training with a feature extractor that directly processes frames from fixed-length videos spanning $T_c$ seconds, corresponding to the context window. The number of frames within this window is denoted as $F_c$. In line with SOTA feature extractors for AS~\cite{xarles2024t}, we use a 2D efficient backbone, RegNetY~\cite{radosavovic2020designing}. This backbone extracts per-frame representations, which are further refined by a transformer encoder-decoder. The encoder, with $l_E$ layers, processes the extracted features, while the decoder, with $l_D$ layers and learnable queries to represent the anticipated actions, captures relevant information from the encoder to anticipate future actions. We further introduce locality biases in the encoder through local self-attention, as shown beneficial in the presented ablations. Finally, each refined query is processed by  the prediction head, which comprises three components: (i) a binary classifier to detect action presence, (ii) a classification module to identify action class, and (iii) a timestamp predictor to estimate the precise temporal position of the anticipated action. Further details of the method are presented in the following sections.

\subsection{Feature extractor}

Following SOTA in Action Spotting, we use RegNetY~\cite{radosavovic2020designing, xarles2024t} as our backbone, known for its efficiency. Additionally, we also incorporate Gate-Shift-Fuse (GSF) modules~\cite{sudhakaran2023gate} in the latter half of the backbone, allowing the extracted features to integrate local spatial modelling before capturing longer temporal dependencies within the transformer encoder. A linear layer is added after the feature extractor to project the representations into $d$ dimensions. As a result, the feature extractor processes the input frame sequence corresponding to the observed context window of $F_c$ frames, $v^n \in \mathbb{R}^{F_c\times H\times W \times 3}$, where $H \times W$ denotes the spatial resolution and produces a sequence of $F_c$, per-frame, $d$-dimensional representations of dimension $F_c\times d$.

\subsection{Transformer encoder-decoder}

The transformer encoder-decoder module processes the features, which primarily contain spatial information along with local temporal information. The objective of the transformer is to capture longer-term temporal dependencies and extract the relevant cues to anticipate actions within the anticipation window. The encoder refines these features, while the decoder transfers the refined information to a set of learnable queries that represent the anticipated actions.

\mysection{Transformer encoder.} It consists of $l_E$ transformer encoder layers, adapted from the original vanilla transformer encoder layers~\cite{vaswani2017attention} to use local self-attention instead of global self-attention, as shown beneficial in \cref{sec:modelAblations}. The frame representations from the feature extractor of dimension $F_c \times d$, pass through $l_E$ encoder layers, each comprising a multi-head local self-attention module with $h$ heads, restricting tokens to attend within a neighborhood of $k$ tokens. Each layer also includes a feed-forward network and layer normalization. Additionally, learnable positional encodings are added to the frame representations at each transformer encoder layer.

\mysection{Transformer decoder.} It consists of $l_D$ vanilla transformer decoder layers. Each layer applies standard multi-head self-attention and cross-attention, both with $h$ heads, followed by a two-layer feed-forward network and layer normalization. The decoder processes a set of $q$ learnable queries representing the anticipated actions. Within each layer, these queries first undergo global self-attention before capturing information from the encoder’s output through the cross-attention mechanism. The resulting refined queries of dimension $q \times d$ are then fed to the prediction head.

\subsection{Prediction head}
\label{subsec:pred_head}

The refined queries, representing $q$ possibly anticipated actions, are fed into the prediction head, which consists of three components: (i) an action detection component, referred to as \textit{actionness}, (ii) an action classification component, and (iii) a timestamp regressor. Each component processes the refined queries in parallel by linearly projecting them through a linear layer. The key differences lie in the output dimensions of 1, $C$, and 1, respectively, and the activation functions, which are sigmoid, softmax, and the identity function, respectively. Thus, the prediction head outputs $\hat{y} = (\hat{y}_d, \hat{y}_c, \hat{y}_t)$, where $\hat{y}_d \in \mathbb{R}^{q \times 1}$, $\hat{y}_c \in \mathbb{R}^{q \times C}$, and $\hat{y}_t \in \mathbb{R}^{q \times 1}$. 
These represent (i) the probability of a query anticipating an arbitrary action, (ii) the confidence for each class, expressed as a probability distribution, and (iii) the absolute temporal position of the action within the anticipation window.

\begin{table*}[t]
  \centering
  \resizebox{\linewidth}{!}{
  \begin{tabular}{p{0.07cm}|lccccccccc|ccccccc}
    \toprule
     \multicolumn{1}{c}{} &  &  &  & \multicolumn{7}{c}{\textbf{mAP@$\delta$ ($T_a=5$)}} & \multicolumn{7}{c}{\textbf{mAP@$\delta$ ($T_a=10$)}} \\
    \cmidrule(lr){5-11} \cmidrule(lr){12-18}
    \multicolumn{1}{c}{} & \textbf{Model} & \textbf{Size} & \textbf{Data} & $\delta=1$ & 2 & 3 & 4 & 5 & $\infty$ & \multicolumn{1}{c}{Avg.} & $\delta=1$ & 2 & 3 & 4 & 5 & $\infty$ & Avg. \\
    \midrule 
    \multirow{4}{*}{\rotatebox{90}{\textbf{\scriptsize ANTICIPATION}}} 
    & FAANTRA & 200MF & SN-BAA & 8.27 & 14.65 & 20.98 & 23.74 & 26.01 & 28.15 & 20.30 & 6.38 & 11.38 & 14.60 & 16.94 & 18.94 & 27.72 & 15.99 \\
    & FAANTRA & 400MF & SN-BAA & 7.83 & 13.41 & 19.49 & 22.58 & 24.50 & 26.65 & 19.08 & 6.03 & 11.37 & 14.30 & 16.28 & 18.17 & 27.80 & 15.66 \\
    & FAANTRA & 200MF & SN-AS + SN-BAA & 9.59 & 17.31 & 23.84 & 27.94 & 30.78 & 32.96 & 23.74 & 5.42 & 10.84 & 16.21 & 18.91 & 21.45 & 32.20 & 17.50 \\
    & FAANTRA & 400MF & SN-AS + SN-BAA & \textbf{9.74} & \textbf{17.47} & \textbf{24.11} & \textbf{28.56} & \textbf{31.13} & \textbf{33.47} & \textbf{24.08} & \textbf{7.32} & \textbf{12.62} & \textbf{18.39} & \textbf{22.08} & \textbf{25.02} & \textbf{33.95} & \textbf{19.90} \vspace{0.12cm}\\
    \midrule
    \multirow{4}{*}{\rotatebox{90}{\textbf{\scriptsize SPOTTING}}} 
    & T-DEED  & 200MF & SN-BAA & 49.70 & 51.91 & 53.40 & 53.62 & 54.44 & 56.06 & 53.19 & 50.66 & 54.17 & 56.04 & 56.44 & 57.28 & 58.25 & 55.47 \\
    & T-DEED  & 400MF & SN-BAA & 52.70 & 55.68 & 57.20 & 57.39 & 57.84 & 59.28 & 56.69 & - & - & - & - & - & - & - \\
    & T-DEED  & 200MF & SN-AS + SN-BAA & 56.72 & 60.89 & 61.95 & 62.54 & 62.95 & 64.02 & 61.51 & \textbf{52.03} & \textbf{54.95} & \textbf{56.05} & \textbf{56.96} & \textbf{57.82} & \textbf{58.93} & \textbf{56.12} \\
    & T-DEED  & 400MF & SN-AS + SN-BAA & \textbf{59.46} & \textbf{61.90} & \textbf{64.34} & \textbf{65.04} & \textbf{65.30} & \textbf{66.99} & \textbf{63.85} & - & - & - & - & - & - & -\\
    \bottomrule
  \end{tabular}
  }
  \caption{FAANTRA's main results on the proposed action anticipation task, evaluated using mAP@$\delta$ metrics from \cref{sec:task}. Results are reported for $T_a=5$ and $T_a=10$ seconds, with models of different feature extractor sizes (in MegaFlops) trained on either SN-BAA alone or jointly with SN-AS. Results from the SOTA Action Spotting method, T-DEED, are also included as an upper bound, under the same variations. Due to limited GPU memory, T-DEED results for $T_a=10$ seconds with a 400MF feature extractor are not reported.}
  \label{tab:mainResults}
\end{table*}

\subsection{Multi-task learning}
\label{sec:training}

The model is trained using a combination of two losses: one for anticipation and one for an auxiliary action segmentation task within the context window.

\mysection{Anticipation loss.} The anticipation loss consists of three components: action detection $\mathcal{L}_D$, action classification $\mathcal{L}_C$, and temporal position prediction $\mathcal{L}_T$. These losses are computed for each query, which are sequentially paired with ground-truth actions in the anticipation window based on their temporal order. When no more ground-truth actions are available, remaining queries are treated as non-actions. For action detection ($\mathcal{L}_D$), we use binary cross-entropy, assigning a positive label to queries paired with ground-truth actions and a negative label otherwise. Then, the action classification loss ($\mathcal{L}_C$) is computed using cross-entropy, with the target being the action class of the paired ground-truth action. Finally, the temporal position loss ($\mathcal{L}_T$) is the mean squared error between the predicted and actual temporal positions of the paired action, with position values scaled by the anticipation window $T_a$ and learned in exponential space, following FUTR~\cite{gong2022future}. For unpaired queries, only the action detection loss is applied. The anticipation loss is then computed as $\mathcal{L}_A = \lambda_D \mathcal{L}_D + \lambda_C \mathcal{L}_C + \lambda_T \mathcal{L}_T$, where $\lambda_D$, $\lambda_C$, and $\lambda_T$ 
weights each loss component.

\mysection{Segmentation loss.} In addition to the anticipation task, the model is trained on an auxiliary action segmentation task, commonly used in anticipation settings~\cite{gong2022future, abu2021long, zhong2023diffant}. This task requires the model to segment actions within the context window, providing direct supervision over frames where actions occur. This helps the model learn the semantics of different action classes, which improves predictions within the anticipation window, as discussed in \cref{sec:modelAblations}. To do so, a linear layer with softmax activation is applied to the output of the transformer encoder, generating a probability vector of size $\mathbb{R}^{F_c \times (C+1)}$, representing the presence of each action class or background at each temporal position. The auxiliary segmentation loss, $\mathcal{L}_S$, is computed using cross-entropy with label dilation~\cite{Hong2022Spotting}, where supervision is applied not only at the exact temporal positions of ground-truth actions but also within a dilation radius around each action, of $r=4$ frames.

\mysection{Multi-task loss.} The final loss function is then computed as $\mathcal{L} = \mathcal{L}_A + \lambda_S \mathcal{L}_S$, where $\lambda_S$ controls the weight of the auxiliary segmentation loss.

\subsection{Inference}

During inference, for each query, we multiply the per-class probabilities ($\hat{y}_c$) with the actionness score ($\hat{y}_d$) to represent the confidence of specific action classes occurring at the predicted temporal position ($\hat{y}_t$), resulting in the final set of anticipated actions.

\section{Results}
\label{sec:results}

In this section, we detail the implementation and training details of our method, the evaluation protocol, and present experimental results.

\subsection{Implementation and training details}

We train our baseline model using a context window of $T_c = 5$ seconds at 6.25 fps (i.e., $F_c = 32$ frames) with a spatial resolution of $448 \times 796$ pixels. The anticipation window is set to either $T_a = 5$ or $T_a = 10$ seconds, as discussed in \cref{sec:taskAblations}. Unless stated otherwise, our model consists of $l_E = 4$ encoder layers, $l_D = 2$ decoder layers, $h = 8$ attention heads, a local encoder with an attention span of $k = 15$ neighboring embeddings, and $q = 8$ learnable queries in the decoder. As feature extractor, we use either RegNetY-200MF or RegNetY-400MF with hidden dimensions of $d = 512$. For $T_a = 10$ experiments, the number of queries is doubled to $q = 16$. All models were trained on an NVIDIA RTX 6000 GPU.

For training, we process untrimmed videos by splitting them into clips of length $T_c$ using a sliding window with a 90\% overlap between consecutive clips, which are randomly sampled. Each clip is linked to annotations for actions occurring within its context window, as well as those in the following $T_a$ seconds (\ie, the anticipation window) for supervision. Models are trained for $30$ epochs with the AdamW~\cite{Loshchilov2019Decoupled} optimizer, using a batch size of $4$ clips, a base learning rate of $1e-4$, $3$ linear warmup epochs, and cosine learning rate decay. The loss weights are set to $\lambda_D=1$, $\lambda_C=1$ $\lambda_T=10$, and $\lambda_S=1$. In the cross-entropy loss, class weights for both auxiliary segmentation and anticipation tasks are set inversely proportional to their frequency. Common data augmentation techniques, including random horizontal flip, Gaussian blur, and color jitter, are applied.

\subsection{Evaluation protocol}

Following standard practices, we train on the train set, use the validation set to prevent overfitting, and evaluate on the test set. Model performance is measured using the mAP variations detailed in \cref{sec:task}.

\subsection{Main results}
In \cref{tab:mainResults}, we present the results of our proposed baseline, FAANTRA, using two variants of the feature extractor (\ie, RegNetY-200MF or RegNetY-400MF). We also report results from training on SN-BAA alone and jointly training on SN-BAA and the original SoccerNet Action Spotting (SN-AS) dataset, which includes $500$ additional games with related action classes. To incorporate SN-AS, we process untrimmed videos similarly to SN-BAA and duplicate the segmentation and anticipation prediction heads, allowing us to leverage both datasets simultaneously, as done by the SoccerNet 2024 Ball Action Spotting challenge winner~\cite{Cioppa2024SoccerNet2024Challenge-arxiv}. Additionally, we include results for the SOTA Ball Action Spotting method, T-DEED, using a context window that matches the anticipation window (\ie, $v^n=w^n$). This serves as an upper bound, where the model has full access to the anticipation window. 

We observe a consistently large performance gap between FAANTRA and the upper bound, particularly evident in low tolerance metrics, highlighting the difficulty of the task --especially its localization component-- and emphasizing the need for further research on anticipation methods in the football domain. Additionally, we observe that a larger feature extractor does not always improve performance when training only on SN-BAA, but it does when incorporating SN-AS. This suggests that the larger backbone benefits from additional data, but is prone to overfitting when limited to SN-BAA for anticipation. Furthermore, adding SN-AS clearly boosts performance, reinforcing the value of this auxiliary dataset. As expected, anticipating actions further in the future is more challenging, as shown by the performance drop when increasing the anticipation window from $T_a=5$ to $T_a=10$ seconds.

\begin{table}[t]
\resizebox{\columnwidth}{!}
{
\begin{tabular}{lccccccc}
\toprule 
& \multicolumn{6}{c}{\textbf{FAANTRA Performance (mAP@$\delta$)}} \\
      \cmidrule(lr){2-7}
\textbf{Action} & $\delta=1$ & 2 & 3 & 4 & 5 & $\infty$ & Avg. \\
\midrule
Pass     & 22.25 & 41.89 & 55.59 & 61.35 & 63.24 & 66.84 & 51.86 \\
Drive    & 28.70 & 47.86 & 57.80 & 63.22 & 65.37 & 70.07 & 55.50 \\
High Pass & 2.30 & 5.32 & 8.33 & 15.92 & 18.86 & 19.60 & 11.72 \\
Header   & 10.30 & 19.87 & 27.11 & 29.69 & 31.18 & 32.19 & 25.05 \\
Out      & 10.70 & 18.26 & 26.86 & 29.35 & 30.82 & 32.38 & 24.74 \\
Throw-in & 2.68 & 9.68 & 15.78 & 28.96 & 37.47 & 45.23 & 23.30 \\
Cross    & 7.84 & 13.22 & 22.51 & 25.31 & 28.98 & 30.26 & 21.35 \\
Ball Player Block & 2.52 & 4.69 & 9.20 & 11.23 & 13.23 & 14.09 & 9.16 \\
Shot     & 4.61 & 7.27 & 9.37 & 11.68 & 12.93 & 14.65 & 10.08 \\
Successful Tackle      & 5.43 & 6.62 & 8.59 & 8.86 & 9.26 & 9.39 & 8.02 \\
\midrule
All      & 9.74 & 17.47 & 24.11 & 28.56 & 31.13 & 33.47 & 24.08  \\
\bottomrule
\end{tabular}
}
\caption{Per-class performance of FAANTRA using RegNetY-400MF, trained on SN-BAA and SN-AS, with results based on the metrics described in \cref{sec:task} for $T_a=5$ seconds.}
\label{tab:perClass}
\end{table}

Additionally, \cref{tab:perClass} presents the per-class results of the best-performing FAANTRA model, using RegNetY-400MF, trained on both SN-BAA and SN-AS, and for $T_a=5$ seconds. We observe that the most frequent actions, such as passes and drives, are better captured by the model. In contrast, actions like headers, ball out, throw-ins or crosses, which often have observable cues before they occur, demonstrate intermediate performance. On the other hand, highly spontaneous actions with limited examples, such as shots, ball player blocks, or successful tackles, are not well captured by the model.

\section{Ablation studies}
\label{sec:ablations}

To assess each component's contribution, we conduct an ablation study by systematically removing or modifying key modules. We evaluate the impact on the test split using the metrics defined in \cref{sec:task}, reporting results averaged over two different seeds. We distinguish between task ablations, which are linked to the task definition; general model ablations, which involve components common to all methods performing the task; and model-specific ablations, which focus on the specific components of our baseline.

\subsection{Task ablations}
\label{sec:taskAblations}

\begin{figure}[t]
    \centering
    \includegraphics[width=1\linewidth]{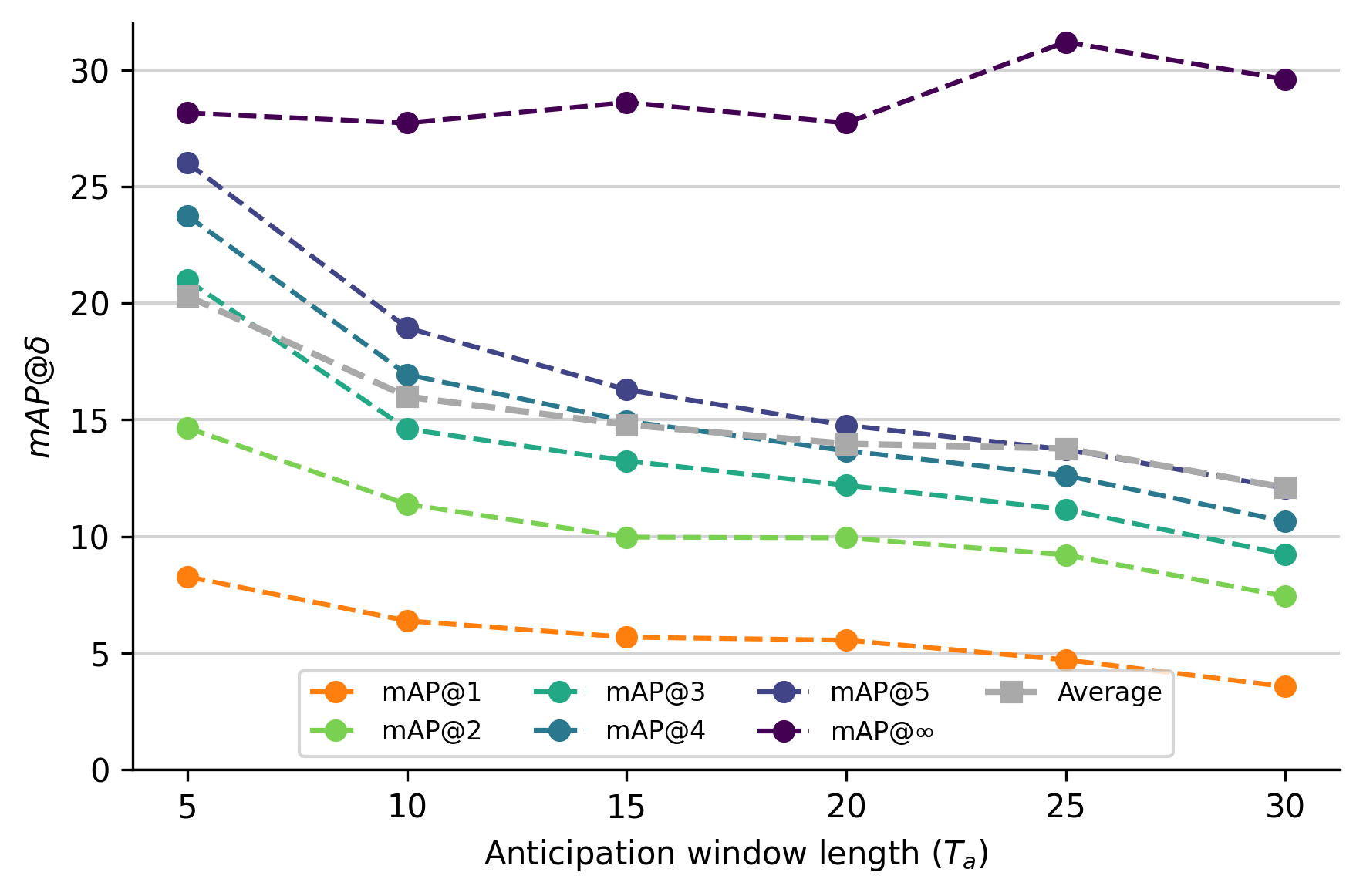}
    \caption{Anticipation window length analysis: Performance evaluation across different mAP@$\delta$ metrics for varying $T_a$ anticipation windows.}
    \label{fig:anticipationWindow}
\end{figure}

In this section, we study the task difficulty with varying anticipation window lengths, $T_a$. As shown in \cref{fig:anticipationWindow}, performance decreases across almost all metrics as the anticipation window increases. This aligns with the nature of the task, where predicting actions further in advance becomes more difficult. The only exception is mAP@$\infty$, which remains stable throughout. This suggests that while the performance related to the proportion of action classes predicted remains unaffected by longer anticipation windows, the challenge lies in localizing those actions. Given the already low scores for smaller windows ($T_a=5$, $T_a=10$), we focus the task on these window sizes, as reflected in the main results (see \cref{tab:mainResults}). Future improvements in action anticipation for these smaller windows will pave the way for tackling the more challenging task of larger anticipation windows. Following ablations are performed using $T_a=5$.

\subsection{General model ablations}
\label{sec:modelAblations}

In this subsection, we analyze the impact of five key components common to any potential action anticipation method: the input frame spatial resolution, temporal resolution, temporal context length, auxiliary action segmentation task within the context window, and prediction head types.

\begin{table}[t]
  \centering
  \resizebox{\columnwidth}{!}{
    \begin{tabular}{llccccccc}
      \toprule
      & & \multicolumn{6}{c}{\textbf{mAP@$\delta$}} \\
      \cmidrule(lr){3-8}
      \multicolumn{2}{l}{\textbf{Experiment}} & $\delta=1$ & 2 & 3 & 4 & 5 & $\infty$ & Avg. \\
      \midrule
      \multicolumn{2}{l}{\underline{FAANTRA}} & 8.27 & 14.65 & 20.98 & 23.74 & 26.01 & 28.15 & 20.30 \\
      \multicolumn{2}{l}{\small{($448\times 796$, 6.25fps, w. AST, Q-Act)}} \\
      \midrule
      (a) & $224\times 398$ & 6.23 & 11.40 & 16.21 & 18.11 & 19.79 & 21.69 & 15.57 \\
      \midrule
      (b)$^\dag$ & 3.125 fps & 4.80 & 12.93 & 15.34 & 18.12 & 19.32 & 21.79 & 15.38 \\
      & 6.25 fps & 6.23 & 11.40 & 16.21 & 18.11 & 19.79 & 21.69 & 15.57 \\
      & 12.5 fps & 5.75 & 11.21 & 16.48 & 18.78 & 19.96 & 22.15 & 15.72 \\
      & 25 fps & 5.29 & 11.56 & 15.47 & 17.85 & 19.79 & 21.47 & 15.24 \\
      \midrule
      (c) & w/o AST & 2.44 & 4.43 & 6.77 & 8.74 & 9.52 & 10.87 & 7.13 \\ 
      \midrule
      (d) & Q-EOS & 6.60 & 12.28 & 17.89 & 19.83 & 21.63 & 23.20 & 16.90 \\ 
      & Q-Bckg & 6.70 & 12.28 & 16.50 & 18.84 & 20.80 & 22.83 & 16.32 \\ 
      & Q-BCE & 5.25 & 9.55 & 13.04 & 15.41 & 16.45 & 18.97 & 13.11 \\ 
      & Q-Hung(t) & 7.38 & 10.28 & 13.01 & 14.97 & 16.52 & 21.36 & 13.92 \\ 
      & Q-Hung(a) & 5.26 & 10.93 & 15.93 & 20.53 & 23.09 & 25.21 & 16.82 \\ 
      & Anchors & 8.39 & 12.09 & 14.71 & 16.72 & 18.95 & 25.15 & 15.84 \\ 
      \bottomrule
    \end{tabular}
  }
  \caption{Ablation studies on general model components from FAANTRA. We ablate the (a) input frame spatial resolution, (b) temporal resolution, (c) auxiliary segmentation task (AST), and (d) prediction head.
  $^\dag$ denotes ablations performed at lower spatial resolution ($224\times398$) due to computational constraints.
  }
  \label{tab:ablationsTask}
\end{table}

\mysection{Spatial resolution.} As shown in \cref{tab:ablationsTask} (a), reducing the spatial resolution by half results in a performance drop of $5$ points on average. This emphasizes the importance of higher resolutions for effectively capturing relevant information within the context window. Since broadcast videos often cover large areas of the field, with key ball-related actions occupying only a small portion of the frame, increased resolution helps capture these important parts of the video more effectively but requires more memory to process.

\mysection{Temporal resolution.} In \cref{tab:ablationsTask} (b), we observe that the optimal frame rate lies between $6.25$ and $12.5$ fps. Lower frame rates likely degrade performance due to missing information from large frame steps, while higher frame rates seem to introduce redundant information without benefiting the method, decreasing the temporal receptive field of the encoder-decoder and increasing computational cost.

\begin{figure}[t]
    \centering
    \includegraphics[width=1\linewidth]{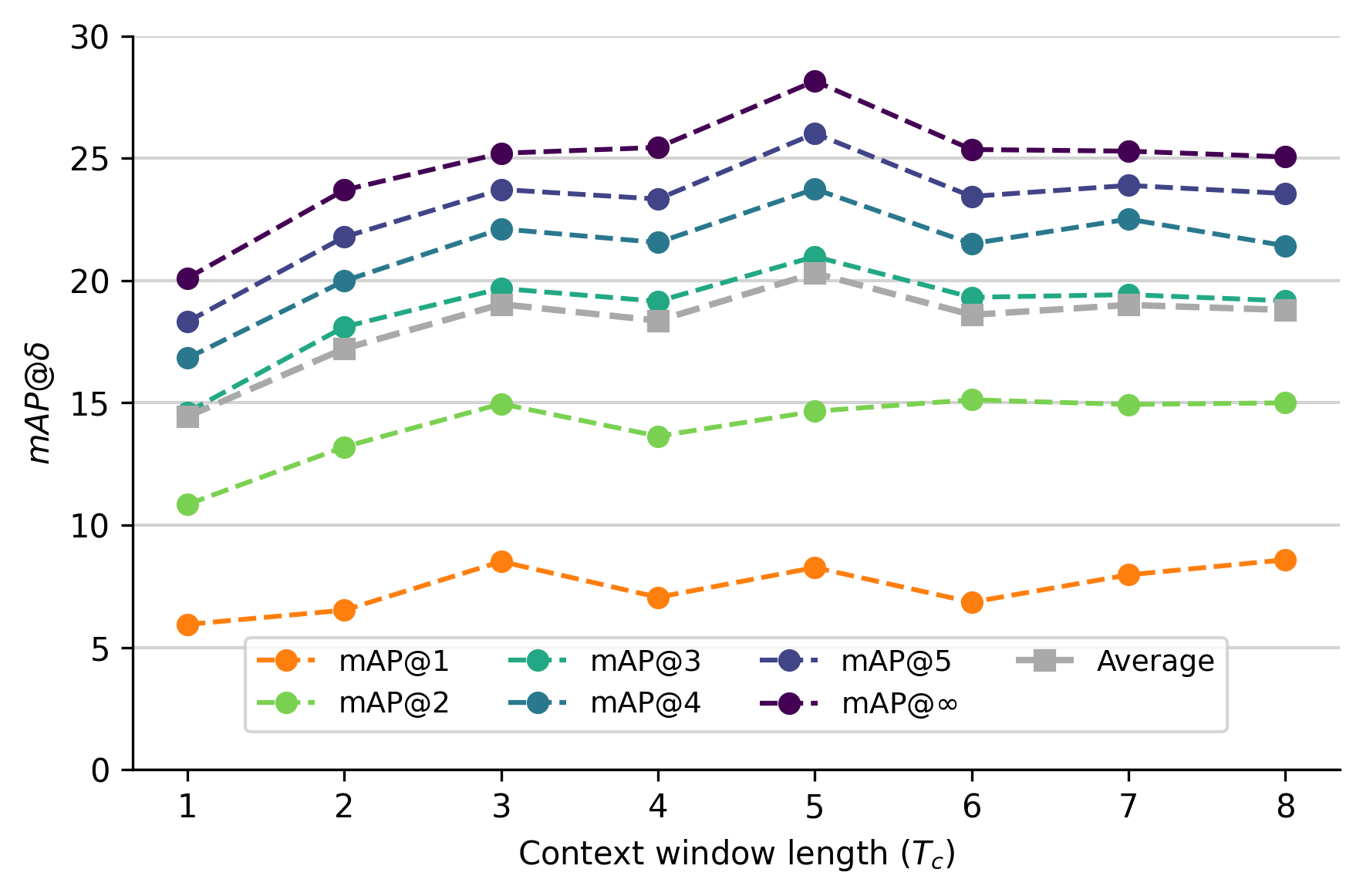}
    \caption{Context window length analysis: Performance evaluation across mAP@$\delta$ metrics for varying $T_c$ context windows.\vspace{-0.15cm}}
    \label{fig:ablationTC}
\end{figure}

\mysection{Temporal context length.} As shown in \cref{fig:ablationTC}, performance initially improves as the context window length increases but plateaus or slightly declines beyond $T_c=5$ seconds. This suggests that, for our proposed baseline, extending the context beyond $5$ seconds offers minimal benefits while increasing computational cost. Given the fast-paced nature of football, information from more distant context appears to add little value for accurately anticipating actions.

\mysection{Auxiliary segmentation task.} \Cref{tab:ablationsTask} (c) highlights the importance of the auxiliary segmentation task in our baseline. Removing this component from FAANTRA results in a substantial performance drop, emphasizing the need for segmentation of observed actions to guide the learning process and capture the semantics of different action classes, which aids in better anticipating future actions.

\mysection{Prediction head.} \Cref{tab:ablationsTask} (d) compares various alternative prediction heads to our baseline, Q-Act, which uses a query-based approach with components for actionness, action class, and temporal position. We evaluate five query-based alternatives and one anchor-based method: Q-EOS, which removes the actionness component and adds an End of Sequence (EoS) action class; Q-Bckg, which replaces EoS with a background class; Q-BCE, which adapts Q-Act by removing the actionness component and using sigmoid and binary cross-entropy for each class instead of softmax and cross-entropy; Q-Hung(t) which pairs ground-truth and predictions based on temporal position using the Hungarian algorithm instead of doing it sequentially; Q-Hung(a), which pairs based on action classes; and Anchors, as in \cite{Soares2022Temporally}, where each query is assigned with a temporal position. Additional details on these alternative prediction heads can be found in the supplementary material.

As shown in \cref{tab:ablationsTask}, Q-Act outperforms Q-EOS, Q-Bckg, and Q-BCE, highlighting the importance of the actionness component within the prediction head. This component simplifies the learning by predicting whether a query contains an action or not independently of its class. For Q-Hung(t) and Q-Hung(a), performance is better when pairing predictions and ground-truths based on action classes rather than temporal position. This makes sense given that the temporal location is more challenging to anticipate. However, this pairing does not outperform sequentially pairing predictions and ground truths. Lastly, using Anchors does not improve performance, likely due to the similar difficulty of action localization, as each query in the Anchors prediction head is itself based on a temporal position.

\subsection{Model-specific ablations}
\label{sec:modelSpecificAblations}

We ablate four architectural components specific to our baseline: (a) encoder layers, (b) decoder layers, (c) attention locality in the transformer encoder, and (d) number of learnable queries. As shown in \cref{tab:ablationsModel}, the optimal number of encoder and decoder layers is $l_E=4$ and $l_D=2$, although  $l_D=4$ results in comparable performance. The model performs better with smaller attention neighborhoods, with $k=7$ and $k=15$ yielding similar results. In contrast, more global attention shows no benefit, highlighting the advantage of starting with more localized attention that can expand to attend to more distant temporal positions in later layers. Finally, $q=8$ queries appears to be optimal, which coincides with the maximum number of actions observed within a $T_a=5$ second anticipation window.

\begin{table}[t]
  \centering
  \resizebox{0.95\columnwidth}{!}{
    \begin{tabular}{llccccccc}
      \toprule
      & & \multicolumn{6}{c}{\textbf{mAP@$\delta$}} \\
      \cmidrule(lr){3-8}
      \multicolumn{2}{l}{\textbf{Experiment}} & $\delta=1$ & 2 & 3 & 4 & 5 & $\infty$ & Avg. \\
      \midrule
      \multicolumn{2}{l}{\underline{FAANTRA}} & 8.27 & 14.65 & 20.98 & 23.74 & 26.01 & 28.15 & 20.30 \\
      \multicolumn{2}{l}{\small{($l_E$=4, $l_D$=2, $k$=15, $q$=8)}} \\
      \midrule
      (a) & $l_E=2$ & 8.17 & 14.24 & 19.61 & 22.45 & 24.38 & 26.80 & 19.28 \\ 
      & $l_E=6$ & 7.12 & 12.69 & 18.05 & 21.25 & 22.74 & 25.09 & 17.82 \\ 
      \midrule
      (b) & $l_D=4$ & 7.86 & 14.23 & 19.76 & 24.46 & 26.87 & 27.17 & 20.06 \\ 
      & $l_D=6$ & 7.81 & 14.12 & 18.97 & 21.43 & 23.96 & 25.90 & 18.70 \\ 
      \midrule
      (c) & $k=7$ & 8.66 & 15.13 & 21.19 & 23.63 & 25.28 & 27.75 & 20.27 \\ 
      & $k=23$ & 8.70 & 14.72 & 19.46 & 21.75 & 23.00 & 25.40 & 18.84 \\ 
      & $k=31$ & 7.77 & 13.55 & 18.34 & 20.98 & 22.49 & 24.96 & 18.01 \\ 
      & Global & 6.33 & 12.28 & 18.42 & 21.18 & 23.05 & 24.96 & 17.70 \\ 
      \midrule
      (d) & $q=6$ & 8.18 & 13.86 & 18.78 & 21.55 & 23.44 & 26.62 & 18.74 \\ 
      & $q=10$ & 8.85 & 14.72 & 19.92 & 23.16 & 24.73 & 26.95 & 19.72 \\ 
      & $q=12$ & 7.27 & 13.20 & 18.07 & 20.63 & 22.05 & 24.48 & 17.62 \\ 
      & $q=14$ & 7.32 & 13.39 & 18.68 & 21.51 & 23.74 & 25.98 & 18.44 \\ 
      \bottomrule
    \end{tabular}
  }
  \caption{Ablation studies of FAANTRA's architectural components: (a) encoder layers $l_E$, (b) decoder layers $l_D$, (c) attention locality in the encoder $k$, and (d) number of learnable queries $q$. \vspace{-0.15cm}
  }
  \label{tab:ablationsModel}
\end{table}

\section{Conclusion}

This work introduced the first video-based action anticipation task in sports, presenting the SoccerNet Ball Action Anticipation dataset. Unlike other anticipation datasets, this task poses an added challenge due to the high spontaneity of football, driven by its adversarial nature. To evaluate the task, we adapted the $mAP@\delta$ metric from action spotting to the anticipation setting. Additionally, we proposed FAANTRA, the first baseline model for this task, and compared its performance to an action spotting upper bound, demonstrating both the feasibility and difficulty of the problem. Finally, we conducted extensive ablations to analyze the impact of different task settings, input configurations, and model architectures, highlighting the importance of high spatial resolution and an auxiliary action segmentation task within the context window to capture the semantics of action classes. By publicly releasing the dataset, baseline, and benchmark, we aim to promote reproducibility and guide future research in football action anticipation. 

\newpage
\mysection{Acknowledgments.} This work has been partially supported by the Spanish project PID2022-136436NB-I00 and by ICREA under the ICREA Academia programme.
This work is supported by the KAUST Center of Excellence for Generative AI under award number 5940.

{
    \small
    \bibliographystyle{ieeenat_fullname}
    \bibliography{abbreviation-short,abbreviation-empty, 
    main}
}

\maketitlesupplementary
\appendix

\section{Extended SN-BAA dataset analysis}

In \cref{tab:datasetStats}, we provide an extended analysis of the SoccerNet Ball Action Anticipation (SN-BAA) dataset. As shown in the table, the original SoccerNet Ball Action Spotting dataset, which has been adapted for the anticipation task, consists of $C=12$ classes. These classes follow a long-tail distribution, with specially evidenced problems in the free-kick and goal classes, where only 21 and 13 examples are observed across the entire dataset respectively. Furthermore, in the test set, these classes appear only 2 and 6 times, making evaluation metrics for these classes highly unstable, as discussed in the main paper. Consequently, these classes are removed from SN-BAA. When analyzing the occurrences of the remaining classes in SN-BAA within an anticipation window of $T_a=5$ seconds, we observe a similar pattern, with passes and drives occurring more frequently, while all other classes have a mean occurrence rate below 0.10.

\begin{table}[h]
\resizebox{\columnwidth}{!}{
\begin{tabular}{lcccccc}
\toprule
\textbf{} & \multicolumn{4}{c}{\textbf{SN-BAS}} & \multicolumn{2}{c}{\textbf{SN-BAA w. $T_a=5$}}  \\
\cmidrule(lr){2-5} \cmidrule(lr){6-7}
\textbf{Action} & Train & Valid. & Test & Total & $\mu$ obs. & Max. obs. \\
\midrule
Pass    & 2679 & 585 & 1721 & 4985   & 0.61  & 6   \\
Drive   & 2297 & 554 & 1449 & 4300   & 0.52  & 4   \\
High Pass     & 465  & 115 & 181 & 761    & 0.09   & 2   \\
Header  & 404 & 127 & 182 & 713    & 0.09   & 5   \\
Out     & 331 & 75 & 145 & 551    & 0.07   & 1   \\
Throw-in     & 213  & 54 & 95 & 362    & 0.04   & 1   \\
Cross   & 177 & 24 & 60 & 261    & 0.03   & 2   \\
Ball Player Block    & 128  & 28 & 67 & 223    & 0.03   & 2    \\
Shot    & 100 & 25 & 44 & 169    & 0.02   & 3    \\
Succesful Tackle   &34   & 12 & 28 & 74     & 0.01   & 2    \\
FK      & 15 & 4 & 2 & 21     & -   & -   \\
Goal    & 6 & 1 & 6 & 13     & -   & -   \\
\midrule
All      & 6849 & 1604 & 3980 & 12433  & 1.52  & 8    \\
\bottomrule
\end{tabular}
}
\caption{Dataset statistics for the SN-BAS action classes, showing the total number of observations for each split, and for SN-BAA, the mean ($\mu$) and maximum number of observations per anticipation window with $T_a=5$ seconds.}
\label{tab:datasetStats}
\vspace{-0.5cm}
\end{table}

\section{Details of prediction heads}
In this section, we provide a more detailed description of the alternative prediction heads analyzed in the ablation studies.

\mysection{Q-EOS.} This approach utilizes the original anticipation head from FUTR. For each query, there are two components: (i) an action classification component, and (ii) a timestamp regressor. The main difference compared to Q-Act is the absence of the action detection component (i.e., actionness), which is replaced by an additional class in the action classification component. This extra class corresponds to an End of Sequence (EoS) class, which is activated when no ground-truth action is paired with the query, signaling the end of prediction generation for subsequent queries. Thus, when an EoS is detected during inference in one query, following queries are discarded.

\mysection{Q-Bckg.} This approach builds upon Q-EOS but replaces the EoS class with a background class. Similar to the EoS class, the background class is activated when no ground-truth action is paired with the query. However, unlike the EoS class, this does not lead to the discard of subsequent queries during inference, and predictions for all queries are considered.

\mysection{Q-BCE.} Similar to Q-EOS and Q-Bckg, this approach omits the action detection component (i.e., actionness) found in Q-Act. Additionally, it modifies the softmax and cross-entropy loss function by using a sigmoid activation function for each class and binary cross-entropy loss, treating the action classes in each query independently. As in Q-BCE predictions for all queries are considered.

\mysection{Q-Hung(t).} This approach adapts Q-Act by modifying the pairing between ground-truth actions and predictions. Instead of sequentially pairing ground truths and predictions, it uses the Hungarian algorithm to pair them based on temporal position, aiming to find the optimal pairing by minimizing the distance between the predicted temporal positions of the queries and the temporal positions of the paired ground-truth actions.

\mysection{Q-Hung(a).} Similar to Q-Hung(t), this approach modifies the pairing between ground-truth actions and predictions. However, in this case, it uses the action classes to perform the pairing. The optimal pairing is determined by minimizing the distance between the predicted scores and the class of the ground-truth action.

\mysection{Anchors.} In this approach, each learnable query is anchored to a temporal window of size $T_a/q$ within the anticipation window $T_a$. During training, each ground-truth token is assigned the first action within its anchor window; otherwise, it is given an actionness value of 0. Only one action is considered within each anchor window, and the temporal position to predict corresponds to the position within the anchor window. During inference, for each prediction, the temporal position is determined by adding the predicted position within the anchor window to the anchor's starting point.

\section{Examples from the dataset}
In this section three frames are shown for each action, a frame before the action (left), the frame of the action label (center), and a frame after the action (right):

\begin{figure*}[t]
    \centering
    \includegraphics[width=1\linewidth]{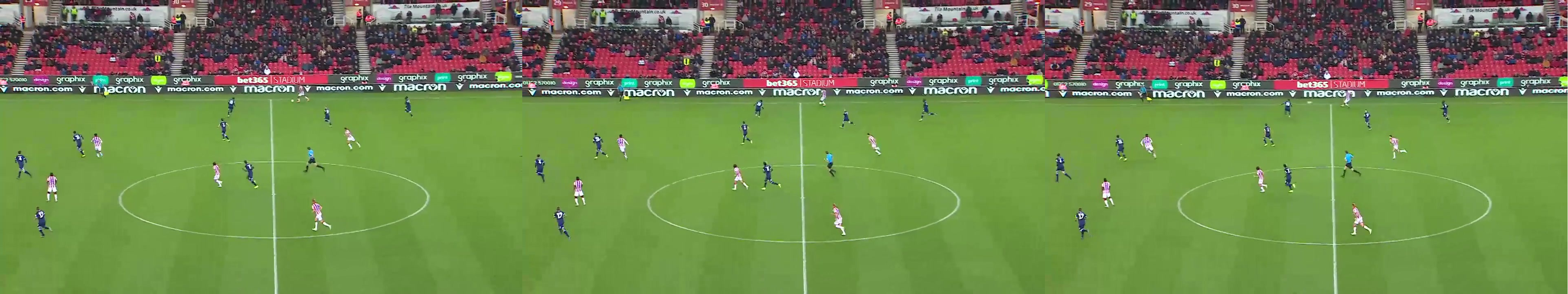}
    \caption{Example of a pass action}
    \label{fig:x-pass}
\end{figure*}

\begin{figure*}[t]
    \centering
    \includegraphics[width=1\linewidth]{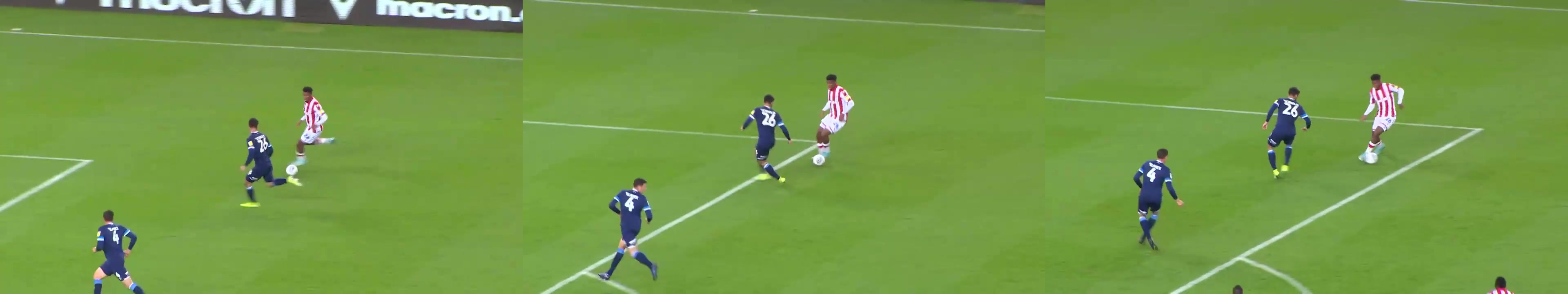}
    \caption{Example of a drive action}
    \label{fig:x-drive}
\end{figure*}

\begin{figure*}[t]
    \centering
    \includegraphics[width=1\linewidth]{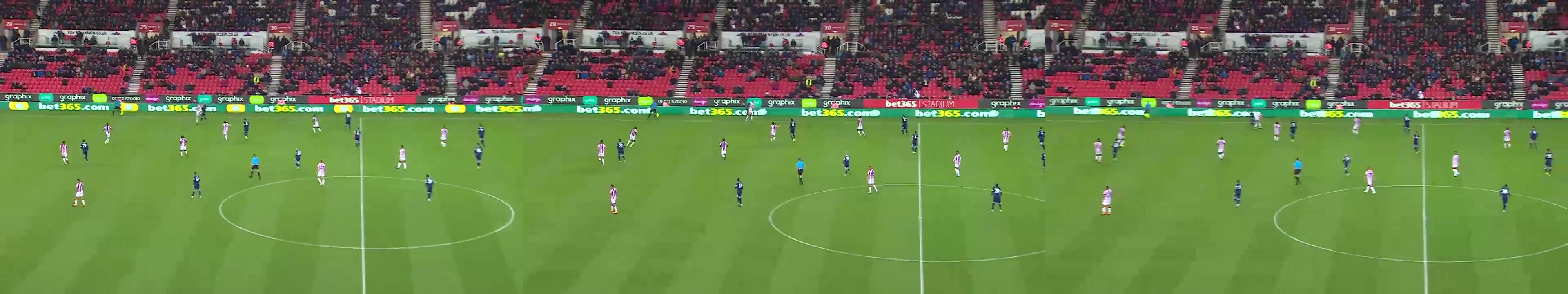}
    \caption{Example of a header action}
    \label{fig:x-header}
\end{figure*}

\begin{figure*}[t]
    \centering
    \includegraphics[width=1\linewidth]{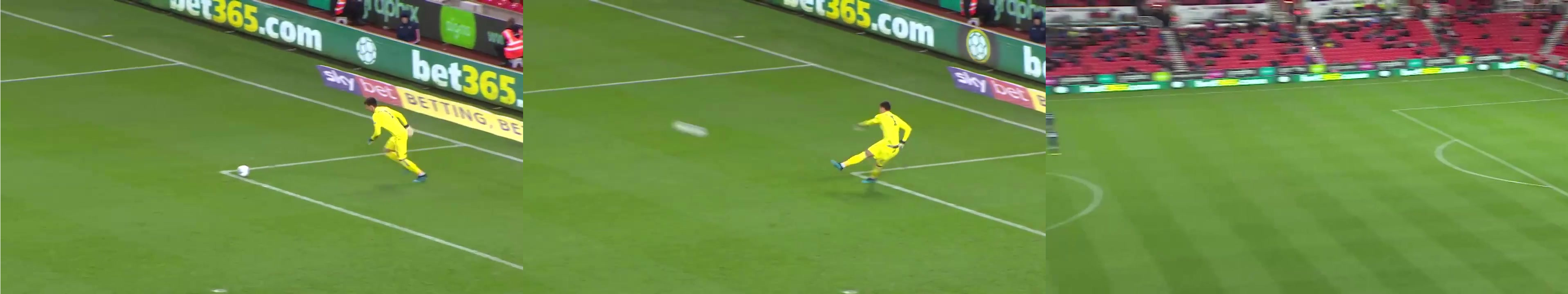}
    \caption{Example of a high pass action}
    \label{fig:ex-high-pass}
\end{figure*}

\begin{figure*}[t]
    \centering
    \includegraphics[width=1\linewidth]{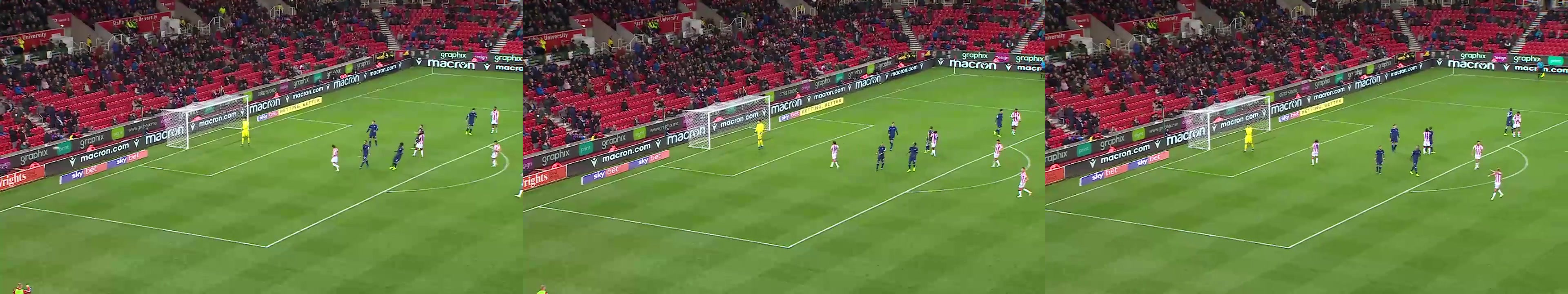}
    \caption{Example of an out action}
    \label{fig:x-out}
\end{figure*}

\begin{figure*}[t]
    \centering
    \includegraphics[width=1\linewidth]{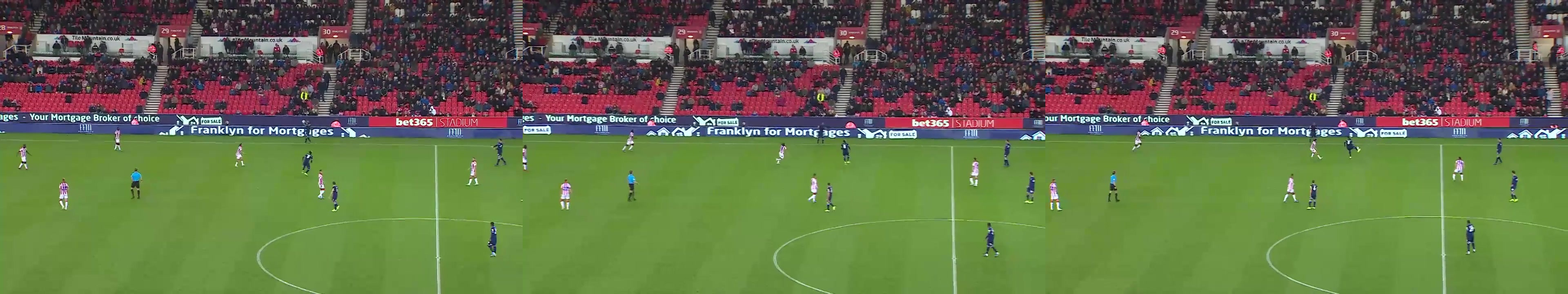}
    \caption{Example of a throw in action}
    \label{fig:x-throw-in}
\end{figure*}

\begin{figure*}[t]
    \centering
    \includegraphics[width=1\linewidth]{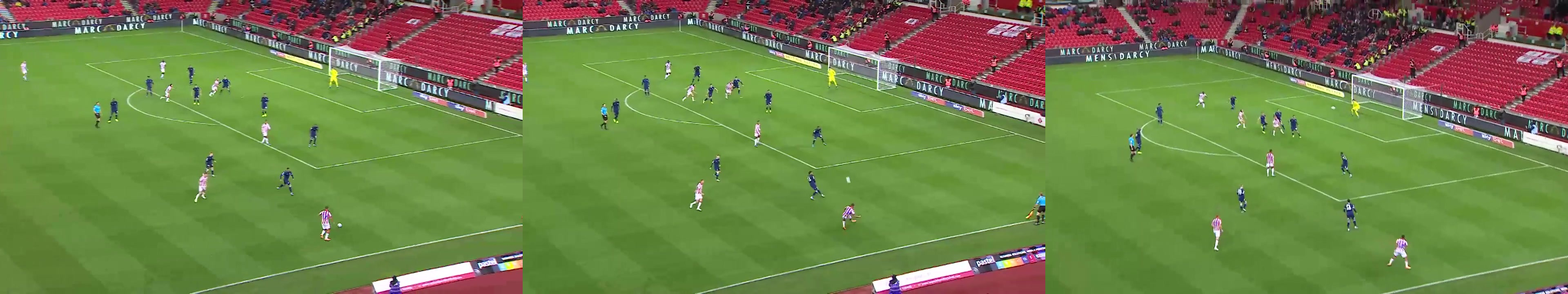}
    \caption{Example of a cross action}
    \label{fig:x-cross}
\end{figure*}

\begin{figure*}[t]
    \centering
    \includegraphics[width=1\linewidth]{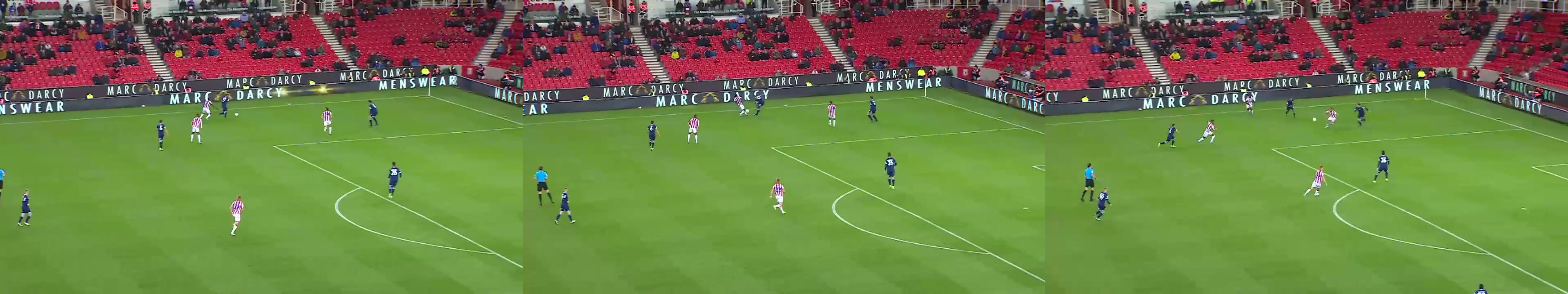}
    \caption{Example of a ball player block action}
    \label{fig:x-ball-player-block}
\end{figure*}

\begin{figure*}[t]
    \centering
    \includegraphics[width=1\linewidth]{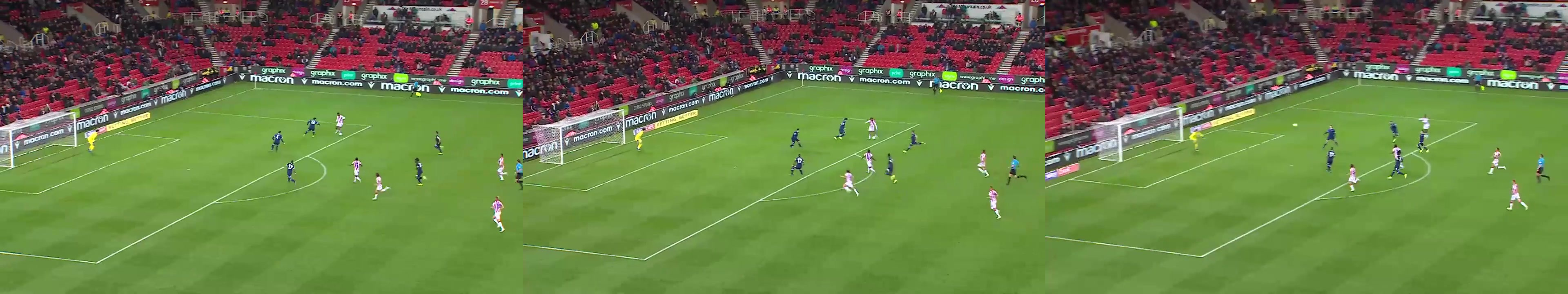}
    \caption{Example of a shot action}
    \label{fig:x-shot}
\end{figure*}

\begin{figure*}[t]
    \centering
    \includegraphics[width=1\linewidth]{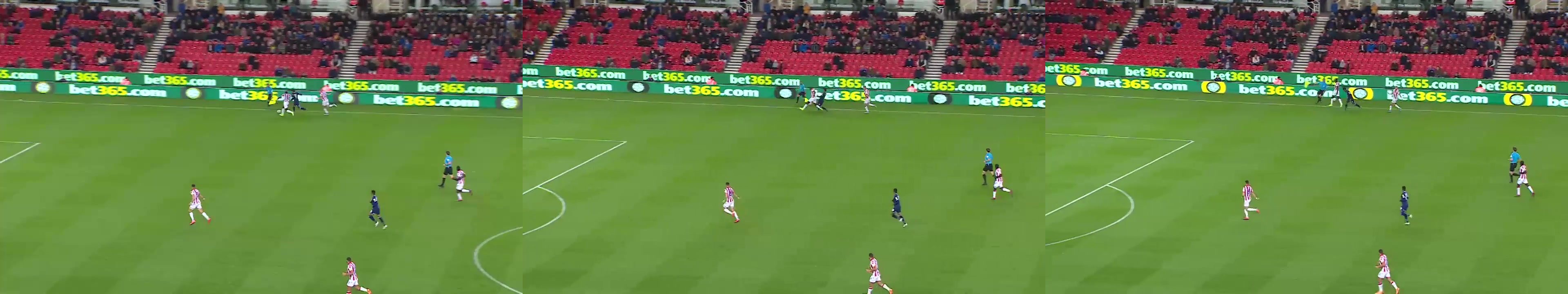}
    \caption{Example of a player successful tackle action}
    \label{fig:x-player-successful-tackle}
\end{figure*}

\end{document}